\documentclass{article}

\usepackage{iclr2027_conference,times}
\usepackage{amsmath,amsfonts,bm}

\def\eqref#1{equation~\ref{#1}}
\def\1{\bm{1}}

\DeclareMathAlphabet{\mathsfit}{\encodingdefault}{\sfdefault}{m}{sl}
\SetMathAlphabet{\mathsfit}{bold}{\encodingdefault}{\sfdefault}{bx}{n}

\usepackage[table]{xcolor}
\usepackage{booktabs}
\usepackage{hyperref}
\usepackage{url}
\usepackage{graphicx}
\usepackage{float}
\usepackage{graphicx}
\usepackage{subcaption}
\usepackage{placeins}
\usepackage{tikz}
\usepackage{capt-of}
\usetikzlibrary{calc}
\usepackage{enumitem}
\usepackage{booktabs}
\usepackage{fontawesome5}

\definecolor{baselineblue}{RGB}{148,167,193}
\definecolor{stackorange}{RGB}{242,163,58}

\newcommand{\bluetext}[1]{\textcolor{baselineblue}{#1}}
\newcommand{\orangetext}[1]{\textcolor{stackorange}{#1}}

\newcommand{\f}{{F$_{1}$}}

\usepackage[most]{tcolorbox}
\usepackage{xcolor}

\definecolor{takeawayorange}{HTML}{CC5908}
\definecolor{takeawayBorder}{HTML}{DDB080}
\definecolor{takeawaybg}{HTML}{FFF4E8}

\newtcolorbox{takeawaybox}{
    colback=takeawaybg,
    colframe=takeawayBorder,
    boxrule=0.9pt,
    arc=1.8mm,
    left=1.6mm,
    right=1.6mm,
    top=0.6mm,
    bottom=0.6mm,
    before skip=8pt,
    after skip=8pt,
}

\newcommand{\shortLine}[1]{{\centering--------------------------------------}\\}

\newcommand{\takeaway}[1]{%
\begin{takeawaybox}
\textcolor{takeawayorange}{\textbf{Takeaway.}}~#1
\end{takeawaybox}
}
\title{Lasting Effects of Abstract\\Pretraining Beyond Perplexity}
\iclrfinalcopy

\begin{document}

\author{%
\begin{tabular}[t]{@{}l@{\hspace{2.5em}}l@{}}
\textbf{Zachary Shinnick}$^{1,*}$ &
\textbf{Hemanth Saratchandran}$^{1}$\\[0.2em]
\textbf{Damien Teney}$^{2}$ &
\textbf{Anton van den Hengel}$^{1,3}$
\end{tabular}\\[1.7em]
{\normalfont\normalsize
$^{1}$Australian Institute for Machine Learning (AIML), Adelaide University}\\[0.10em]
{\normalfont\normalsize
$^{2}$Idiap Research Institute \qquad $^{3}$Metacognition AI}\\[0.3em]
{\normalfont\small
$^{*}$Correspondence: \texttt{zachary.shinnick@adelaide.edu.au}}%
}

\maketitle
\vspace{-4pt} % Z: Uncomment
\begin{abstract}
Language models are typically pretrained from random initialization. Recent work challenges this convention, showing that a brief warm-up on abstract, algorithmically generated data can provide a better starting point for subsequent learning of natural language.
In this paper, we show that in small language models,
such a warm-up improves
specific capabilities
%the acquisition of
%downstream capabilities
%in ways
that are not reflected in language-modeling perplexity.
Our warm-up uses an abstract stack-manipulation task
that requires compositional and state-tracking capabilities.
%in which the model processes sequences of push and pop operations and predict the remaining contents.
Allocating as little as 1\% of pretraining tokens
to this data
%as many tokens as the language-pretraining phase that follows, it 
improves multi-hop question answering %after fine-tuning
by up to 3.9~\f\,points on \textsc{MuSiQue},
with additional gains on \textsc{HotpotQA} and \textsc{2WikiMultihopQA}
despite comparable language-modeling perplexity.
Controlled experiments % further
show that the warm-up
substantially
accelerates the acquisition of deeper reasoning chains.
We also explore what drives this transfer.
First, the %precise
structure of the data matters:
replacing the stack task with a queue fails to produce the same gains.
%stack's last-in-first-out rule with a matched first-in-first-out rule eliminates the downstream advantage.
Second, the gains are specific:
performance improves
on sequential reasoning chains,
%when later reasoning depends on earlier steps,
with no consistent benefit on tasks
that combine or compare independent facts.
%transfer is selective:
%gains concentrate on dependent reasoning chains, with no consistent benefit for tasks that merge or compare independent lookups.
Third, timing matters:
mixing abstract data with natural language is far less effective than
an initial dedicated phase,
and %using it
exposure
\emph{after} pretraining completely removes the benefits.
%exposure before language pretraining produces the largest gains, during pretraining yields smaller gains, and afterwards has no improvement. 
%And 
%Moreover,
The early advantage %also 
persists through billions of subsequent language tokens.
% These results show that early training on abstract %algorithmic
% data can reliably shape
% the capabilities that language models later acquire. 
These results show that early abstract training can reliably shape
the capabilities language models later acquire.\\[5pt]
\faGlobe\enspace\textbf{Project page:}\enspace
{\footnotesize
\href{https://zlshinnick.github.io/beyond-perplexity/}{%
  \nolinkurl{zlshinnick.github.io/beyond-perplexity/}}
}
\end{abstract}

\vspace{-9pt}
\section{Introduction}
%\vspace{-1pt}
Recent studies show that brief training on abstract, algorithmically generated data can improve how transformers subsequently learn language, code, mathematics, and images%
~\citep{jiang2026proceduralpretraining,lee2026training,papadimitriou2023injecting,shinnick2025can}.
This approach, known as \emph{procedural pretraining}, exposes models to data generated by simple algorithms such as formal languages or cellular automata, prior to standard pretraining (also called \emph{pre-pretraining} by \citealt{hu2025between}).
The benefits include faster convergence and improved downstream capabilities~\citep{jiang2026proceduralpretraining,lee2026training}.
Yet it remains unclear which properties of the abstract data matter, which specific capabilities improve, and what drives these benefits.

This paper studies how abstract pretraining shapes the subsequent acquisition of capabilities in small language models,
in particular for multi-hop reasoning.
We show that brief initial training on an abstract memory-manipulation task improves multi-hop question answering and accelerates the acquisition of reasoning skills over long chains.
Importantly, these benefits are not always reflected
in language-modeling perplexity.
This suggests that abstract pretraining shapes the inductive biases for generalization, rather than simply making the language-modeling optimization easier.
It also implies that selecting abstract data based on perplexity alone can miss important differences in the capabilities they enable.

\begin{figure}
    \centering
    \includegraphics[width=0.99\linewidth]{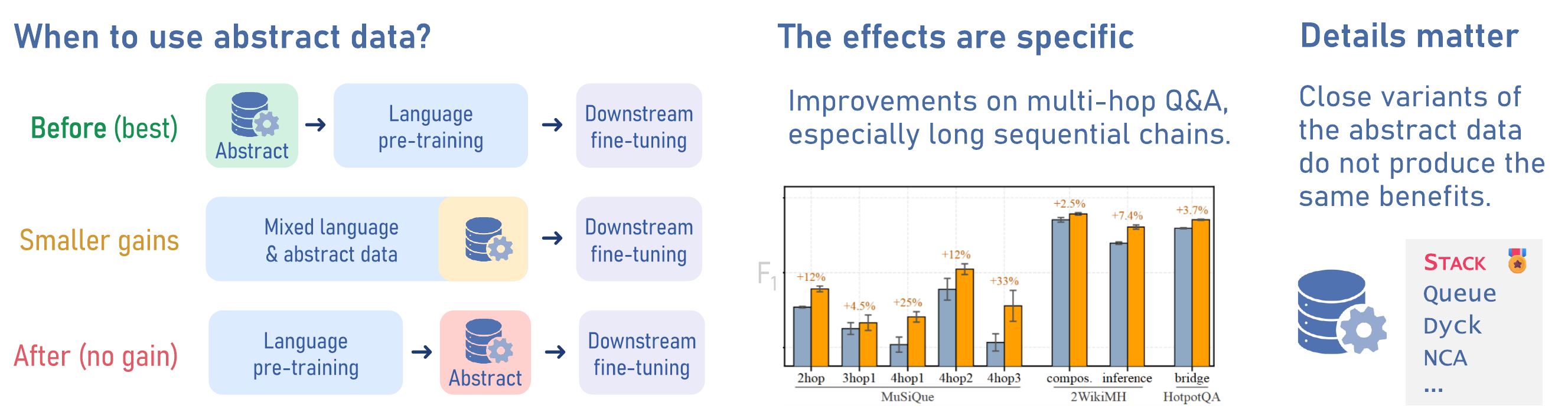}
    %\vspace{1pt}
    \caption{%
    We identify three practical principles for using abstract data to pretrain language models.
    \textbf{(Left)}~It is most effective %when used
    in an initial dedicated phase.
    \textbf{(Middle)}~The benefits are not always apparent in language-modeling perplexity, but they show up more clearly in evaluations of specific capabilities such as multi-hop reasoning.
    \textbf{(Right)}~The choice of abstract data matters: the benefits depend on the model learning specific structures rather than %getting
    a generic headstart on the optimization.
    }
    \label{fig:teaser}
    \vspace{-8pt}
\end{figure}

\textbf{Overview of our study.}
We conduct controlled experiments to identify which capabilities benefit from abstract training, which properties of the abstract data drive the transfer, and how the timing of exposure affects it.
We evaluate multi-hop question answering on \textsc{MuSiQue}, \textsc{HotpotQA}, and \textsc{2WikiMultiHopQA}~\citep{trivedi2022musique,yang2018hotpotqa,ho2020constructing} and use a controlled synthetic task (\textsc{DEPO}, \citep{allen2026physics}) to measure the acquisition of increasingly long reasoning chains.
We also vary the structure of the abstract task
and its timing relative to language pretraining.
This allows us to isolate the specific capabilities that transfer and determine whether the same abstract data must be presented before, during, or after language pretraining.

We focus on a specific form of abstract data, referred to as \textsc{Stack}, that simulates sequences of push and pop operations on a memory.
Next-token prediction on this data requires tracking the state of the memory through nested, compositional sequences.
Prior work found it particularly effective for procedural pretraining of LLMs for code and natural language~\citep{jiang2026proceduralpretraining}.
Its simple structure also allows us to systematically manipulate the abstract computation and test which aspects of the learned inductive bias transfer to downstream capabilities.

Our contributions and main findings are summarized as follows.
\begin{enumerate}[leftmargin=15pt, itemsep=2.0pt, parsep=1.5pt, topsep=-2.5pt]
\item \textbf{Identifying capabilities that improve with abstract pretraining.}
We find clear improvements in multi-hop reasoning on Q\&A benchmarks
(\textsc{MuSiQue}, \textsc{HotpotQA}, and \textsc{2WikiMultiHopQA}),
specifically on reasoning chains in which later steps depend on earlier ones, rather than on tasks that merge or compare independent facts.
Using \textsc{DEPO}, we further find faster acquisition of longer reasoning chains (Section~\ref{sec:evaluating}).
These effects are not consistently coupled with improvements in language-modeling perplexity.

\item \textbf{Identifying properties of the abstract data that determine transfer.}
We construct variants of the \textsc{Stack} task that fail to reproduce its benefits (Section~\ref{sec:characterizing}).
In particular, a similar queue-manipulation task achieves similar perplexity but does not produce the same downstream improvements.
Thus, the choice of abstract data can affect generalization independently of its effect on language modeling.

\item \textbf{Identifying when abstract training must occur.}
We find that early exposure is critical: abstract training throughout or after language pretraining does not produce the same benefits (Section~\ref{sec:training_trajectory}).
The persistence of the early advantage through billions of subsequent language tokens shows that abstract training can meaningfully influence the whole training trajectory.
%The need for early exposure suggest a meaning impact on the whole training trajectory.
%And the fact that the benefits cannot be obtained through post-training suggests that abstract data instills capabilities that our baseline models intrinsically lack.
\end{enumerate}
\vspace{2pt}

Our results identify three factors that affect
whether abstract data produces measurable benefits:
the capability being evaluated,
the structure of the abstract data,
and its position in the training pipeline.
Our experiments are conducted at a small scale (SmolLM-135M and 360M) but they complement prior studies showing that
the benefits of abstract training extend to
%larger models and longer language-training regimes~\citep{jiang2026proceduralpretraining,cheng2026logic}.
larger models
(1.3B parameters in \citealt{jiang2026proceduralpretraining})
and longer training
($\sim400$ language tokens/parameter in \citealt{cheng2026logic}).

\vspace{-4pt}
\section{Related Work}
\vspace{-4pt}
\textbf{Pretraining on abstract data.}
Data from formal languages is a useful tool
in linguistics and NLP
for controlled studies of language acquisition
~\citep{chiang2022transferability,goodale2025meta,mccoy2023modeling,papadimitriou2023injecting,ri2022pretraining,hu2025between}.
Abstract data was proposed recently
as a practical means to improve the training of large language models (LLMs)
~\citep{lindemann2024sip,wu2022insights,wu2021lime,zhang2024intelligence,bloem2025universalpretrainingiteratedrandom}.
This data is used in a short initial training phase,
and the benefits include faster subsequent training
\citep{jiang2026proceduralpretraining,lee2026training},
robustness to noisy data~\citep{guo2026synthetic},
and improvements in downstream capabilities
%on benchmarks %downstream capabilities
%that do not necessarily manifest in better perplexity 
%on the pretraining data
\citep{cheng2026logic,veitsman2026structural}.
Improvements on benchmarks are not necessarily tied to
better pretraining perplexity, which is well known for LLMs in general~\citep{liu2023same,velivckovic2026perplexity}.
%This paper explores
We explore this in more detail,
identifying properties of the data responsible for the improvements
and the effect on specific capabilities such as multi-hop reasoning.

\textbf{Training order in language-model pretraining.}
Models learn different skills and properties of the data
in a predictable order \citet{achille2019critical,belrose2024neural,michaelov2026language}
which can be controlled
with the order of the data ~\citep{bengio2009curriculum}.
%The order of the training data can affect learning,
%as shown in work on curriculum learning and critical learning periods~\citep{bengio2009curriculum,achille2019critical}. In language models, 
Reasoning data during pretraining provides benefits that
%later
fine-tuning does not fully reproduce~\citep{akter2025front}.
Most closely related to our work, \citet{cheng2026logic} show that initial training on formal derivations affects subsequent skill acquisition and model representations. However, this and other studies with abstract data do not establish whether a dedicated initial training phase is necessary. We address this gap by directly comparing exposure to abstract data before, during, and after language pretraining, and testing whether the effects %of early exposure
persist through subsequent training.

\textbf{Multi-hop reasoning} is the skill required to
combine multiple pieces of evidence
through dependent steps. It is challenging for language models
and dedicated question-answering benchmarks
are designed such that the answers cannot be determined by retrieving individual facts~(e.g.\ 
HotpotQA, 2WikiMultiHopQA, MuSiQue; \citealt{yang2018hotpotqa,ho2020constructing,trivedi2022musique}). Existing solutions include
context repetition~\citep{yu2025unleashing},
interleaved retrieval and reasoning~\citep{trivedi2023interleaving},
and RL fine-tuning~\citep{kabra2026learning}.
%Models can also remain sensitive to the ordering of supporting passages even when irrelevant context is removed~\citep{yu2025unleashing}.
In comparison, we keep the architecture and fine-tuning protocol fixed,
and show that multi-hop reasoning can be improved
%implicitly
by shaping the model's inductive biases
through an initial training on generic abstract data.

%We evaluate this transfer primarily on established multi-hop benchmarks and complement them with an adaptation of DEPO~\citep{allen2026physics}, which measures how quickly models learn increasingly long chains of relations.

\vspace{-4pt}
\section{Methods}
\label{sec:methods}
\vspace{-3pt}

%\subsection{Overall setup}
\label{sec:experimental_setup}

Our setup follows prior work on abstract data (see e.g.\ \citealt{jiang2026proceduralpretraining,lee2026training}).
We train a language model first
on a small amount of abstract data,
then proceed
with standard phases
such as language pretraining
and supervised fine-tuning (SFT) on a target task.
%The model thus undergoes three phases:

\vspace{2pt}
\centerline{
\footnotesize
\setlength{\tabcolsep}{1pt}
\renewcommand{\arraystretch}{1.3}
\begin{tabular}{ccccc}
&& \multicolumn{3}{c}{\textit{Baseline}} \\[-2.7pt]
\cmidrule{3-5}
{\scriptsize(1)}~~\textbf{Training on abstract data} & ~~$\rightarrow$~ &
{\scriptsize(2)}~~\textbf{Standard language pretraining} & $\rightarrow$~~ &
{\scriptsize(3)}~~\textbf{SFT on target task} %\& Evaluation
\\[-2.7pt]
\scriptsize
\textsc{Stack}, \textsc{Queue}, \textsc{Nca}, etc. &&
\scriptsize
\textsc{FineWeb-Edu} or \textsc{TinyStories} &&
\scriptsize
~~\textsc{HotpotQA}, \textsc{Depo}, etc.
\\[-4.1pt]
\scriptsize
Typically $\sim$80M tokens
&&
\scriptsize
E.g.\ 7.2B tokens
(20 tokens/param.) for SmolLM-360M
&&
\scriptsize
Budget varies across datasets
\end{tabular}
}
%\vspace{1pt}

The model is evaluated for 
language-modeling perplexity after (2)
%validation loss at matched language-token
and for downstream performance on the target task after (3).
%counts and measure downstream performance after fine-tuning.
The main baseline is a model that undergoes
only phases (2) and (3).
The first phase on abstract data can therefore be seen as an alternative to standard random initialization that instills
better %generic
inductive biases in the model.
We match the token budget of phases (2,3)
across experiments
since the additional data in the abstract phase
only adds a very small amount of tokens
($\sim$1\% of the pretraining budget,
see Table~\ref{tab:procedural_task_configs} in the appendix).

%We will see in Section~\ref{sec:???} that the benefits far outweigh the expected nudge in performance from a similar increase in language tokens in phase (2). 

%\damien{The original statement "\textit{We match the token budgets.}" was ambiguous!!! (with/without abstract data: do you match the *total* budget? If not people will say it's cheating. You then need to explain why it's reasonable not to include the abstract data in the budget; e.g. cite our paper that evaluated multiple settings). Also mention that abstract data is about 1\% additional initial training. Edit: I tried to fix it myself. I'm still not sure what is the justification for not matching the *total token budget* (the fairer setting).} \zach{I think the justification is we are manipulating the inductive biases in phase 1, so we are comparing the setups in phase 2 and 3. Also we could refer to figure 12 where we can see reducing the NL exposure for the stack model, it reaches the same performance with roughly 30-40\% less data than then baseline.}

We additionally want to evaluate whether the abstract data
could be used at other points in this pipeline. We therefore evaluate alternative settings (see Figure~\ref{fig:teaser})
where the abstract data is either interspersed with the language data,
or used after language (swapping (1) and (2) above).
%i.e.\ interverting (1) and (2) above.

%TODO: why you want to vary this setup. What alternatives you evaluate (Figure~\ref{fig:teaser})

%Old text:
%The language-only baseline omits the first stage and begins language pretraining from random initialization. Within each setting, we match language-pretraining data, token budgets, and optimization settings, as well as downstream training protocols. The procedural condition receives roughly 1\% additional initial training.
%We compare language-modeling validation loss at matched language-token
%counts and measure downstream performance after fine-tuning.
%Natural-language QA measures performance after a fixed fine-tuning
%budget, while controlled learning curves track learning across varying multi-hop composition lengths. We also conduct controlled experiments that omit language pretraining to measure direct transfer from abstract tasks.

%------------------------------------------------------------
\vspace{-3pt}
\subsection{Models and Language Pretraining}
\label{sec:natural_language_setup}
\vspace{-2pt}

%\textbf{Architectures}.
Our main experiments use the SmolLM-135/360M
architectures%
%and tokenizer
~\citep{allal2024SmolLM}
and \textsc{FineWeb-Edu}
\citep{penedo2024fineweb}
as the source of language data,
with a budget of 20 language
tokens/parameter~\citep{hoffmann2022training}.
Our abstract data (described below)
uses the same vocabulary size as the SmolLM tokenizer.
The token embeddings and output head
are thus trained and transferred across phases.
A few %of our
experiments use
\textsc{TinyStories}~\citep{eldan2023tinystories}
as language data.
%as an alternative source of language data.

%\textbf{Hyperparameters.}
We use %the %training recipe and
hyperparameters
from the SmolLM codebase
that we adapt for the Muon optimizer
because it performs significantly better as a baseline.
We re-tune the 
learning rates and weight decay
by grid search with the SmolLM-135M baseline
with full Chinchilla-optimal runs
(2.69B tokens; 
details in 
Appendix~\ref{app:nl_procedural_training}
and~\ref{app:nl_standard_pretraining})
and use these hyperparameters throughout our experiments.
We do not re-tune them
when using abstract data,
so further gains might be possible
%Our results are reported over three to nine seeds
(details in Appendix~\ref{app:natural_language_evaluation}). %\zach{Could add sentence here and refer too appendix about setup for the controlled experiments.}

%------------------------------------------------------------
\vspace{-2pt}
\subsection{Abstract data}
\label{sec:procedural_pretraining}
\vspace{-2pt}

For abstract training,
we optimize the model
for next-token prediction
on one of these types of data.

\textbf{Memory manipulation (\textsc{Stack}).}
This data is generated as random sequences of
push and pop operations on a simulated stack memory (last-in, first-out).
Most operations are arbitrary (at any point,
valid continuations are to either push any new token,
or pop one if the stack is not empty).
But each sequence ends with the actual contents of the memory
according to the preceding operations (see 
examples in Table~\ref{tab:stackExample} and
Appendix~\ref{app:procedural_tasks}).
Successful prediction therefore requires
tracking the state of the memory over the whole sequence.
This data proved particularly effective in prior work
\citep{jiang2026proceduralpretraining}
and in our initial experiments.
It is therefore the focus of our experiments.

\setlength{\fboxsep}{2pt}
\newcommand{\phantomheight}{\hspace{-1.25em}\phantom{p(}} % Normalize height of boxes
\newcommand{\token}[1]{%
  \fcolorbox{gray!95}{white}{\scriptsize\texttt{#1\phantomheight}}~~%
}
\newcommand{\lossToken}[1]{%
  \fcolorbox{gray!85}{yellow!25}{\scriptsize\texttt{#1\phantomheight}}~~%
}

\textbf{Variants of \textsc{Stack}.}
In the first variant (\textsc{Queue}),
we replace the stack with a queue (first-in, first-out).
In the second variant (\textsc{StackRand}),
we randomly permute the final contents
of the stack.
%in each sequence.
These  allow evaluating minor variations
in the underlying structure of the data
while keeping the surface statistics similar.

% \damien{I added names to the variants. Make sure you use these names later in the text.}

% \damien{StackRand is a bit weird, I understand now that it's actually making the task more difficult because the model will never get the correct answer (since it's random). A better choice
% (trickier to implement)
% would have been to count as correct *any* of the possible correct predictions (not just 1 arbitrary permutation).}

% \damien{I wrote examples above using separator tokens. I think you did not include them in your code, which doesn't really make sense (the EOS should be there to teach the model when to stop, and SEP is necessary to tell the model when to start answering).
% I suggest keeping the correct version (the one I wrote)
% even if it doesn't exactly match your code.
% Writing your version will be confusing and there's no good reason to explain why you did it that way.}
% \zach{Im pretty sure there is a seperator, will double check.}

\textbf{Alternative tasks.}
We also evaluate others forms of abstract data proposed in prior work,
generated with simple algorithms like 
\textsc{Sorting} and \textsc{Set}~\citep{jiang2026proceduralpretraining},
neural cellular
automata (\textsc{Nca},~\citealt{lee2026training}),
and formal languages~\citep{hu2025between}.
See details in
Appendix~\ref{app:procedural_tasks}--~\ref{app:controlled_procedural_training}.

\begin{table}[h]
\centering
\vspace{-4pt}
\caption{Examples of abstract data. Boxes represent tokens.
The loss is computed on colored ones.}
\vspace{-6pt}
%\caption{Examples of different types of abstract data. Boxes represent individual tokens. The training loss for next-token prediction is computed only on colored ones.}
\label{tab:stackExample}
%\footnotesize
\scriptsize
\setlength{\tabcolsep}{3pt}
\renewcommand{\arraystretch}{1.0}
\begin{tabular}{ll}
\toprule
{Type of abstract data}~~~~~~ & {Example sequences}\\
\midrule
\textsc{Stack}&
\token{push(A)}\token{push(B)}\token{pop()}\token{push(C)}%
\token{<SEP>}\lossToken{A}\lossToken{C}\lossToken{<EOS>}\\[2pt]
\textsc{Queue}&
\token{push(A)}\token{push(B)}\token{pop()}\token{push(C)}% 
\token{<SEP>}\lossToken{B}\lossToken{C}\lossToken{<EOS>}\\[2pt]
\textsc{StackRand}&
\token{push(A)}\token{push(B)}\token{pop()}\token{push(C)}%
\token{<SEP>}\lossToken{C}\lossToken{A}\lossToken{<EOS>}\\[2pt]
\textsc{Sort}&
\token{Z}\token{C}\token{B}\token{T}\token{A}%
\token{<SEP>}%
\lossToken{A}\lossToken{B}\lossToken{C}\lossToken{T}\lossToken{Z}\lossToken{<EOS>}\\[2pt]
\bottomrule
\end{tabular}
\vspace{-10pt}
\end{table}

% \damien{I don't see Sorting/Set in the experiments, but I see Union Dyck ?!} \zach{Full search over procedural tasks is in the appendix (Figure 14).}

%------------------------------------------------------------
\vspace{-2pt}
\subsection{Target Data and Evaluation}
\label{sec:methods_eval}
\vspace{-2pt}

We use three question-answering %(Q\&A)
benchmarks designed
to evaluate multi-hop reasoning:
%We evaluate transfer to natural-language multi-hop question answering and a controlled synthetic task for relational reasoning:

\begin{itemize}[leftmargin=15pt, itemsep=3pt, parsep=1pt, topsep=-2.5pt]
\item \textbf{\textsc{MuSiQue}}~\citep{trivedi2022musique} evaluates multi-hop reasoning through questions composed from dependent single-queries, with 2--4 reasoning steps that prevent the use of disconnected shortcuts. We use the answerable split with gold supporting paragraphs.

\item \textbf{\textsc{HotpotQA}}~\citep{yang2018hotpotqa} evaluates multi-hop reasoning over multiple supporting documents, including bridge and factoid-comparison questions.

\item \textbf{\textsc{2WikiMultihopQA}}~\citep{ho2020constructing} evaluates multi-hop reasoning %on Wikipedia
using compositional and comparison questions
derived from Wikidata relations.
\end{itemize}
We follow
the protocol from \citet{trivedi2022musique}
and fine-tune the model for one epoch
on 20k training questions
(details in Appendix~\ref{app:natural_language_evaluation}).
We report the mean token-level \f~in our main results.
Other metrics are reported in Appendix~\ref{app:additional-results}.
We also use a synthetic task for controlled evaluations:
%with a minimal architecture (Appendix \ref{app:additional-results}) for relational reasoning:
\begin{itemize}[leftmargin=15pt, itemsep=3pt, parsep=1pt, topsep=-2.5pt]
\item \textbf{\textsc{DEPO}}~\citep{allen2026physics}
evaluates multi-hop reasoning
%over facts in the context
%with queries
of controlled complexity.
%whether a model can perform relational reasoning of controlled complexity.
Each sequence describes a directed cycle of nodes
with the list of its connected pairs in random order.
This is followed by a ``query'' that specifies a
node and number of hops $k$.
The answer is then the node reached $k$ steps down the chain.
The model must therefore dereference $k$ pairs from the context
in the right order to identify the correct answer.
We use the code from \citet{allen2026physics}
to generate data for fine-tuning and evaluation
with a difficulty %level$
adjusted to be barely solvable by our baseline models,
namely 10 nodes per cycle
%followed by 10 queries,
and $k\in\{$1,2,3,4$\}$
(details in Appendix~\ref{app:depo}).
%for each query, with loss applied only to answer tokens.
%We measure teacher-forced answer accuracy on fixed evaluation sets for each hop count.
%Task encoding, optimization, and evaluation details appear in
\end{itemize}

% \damien{You re-added
% "\textit{We also use a synthetic task with a minimal architecture}".
% I removed it. Like I said, this has no place here (3.3 TARGET DATA AND EVALUATION).
% }

\begin{comment}
PUT ALL OF THIS SOMEWHERE ELSE

\subsection{Controlled Reasoning Experiments}
\label{sec:controlled_setup}
We use a four-layer GPT-2 transformer with hidden dimension 512 and four attention heads.
Models transfer either directly from procedural training to the
downstream task or through intervening language pretraining on
\textsc{FineWeb-Edu}~\citep{penedo2024fineweb} or
\textsc{TinyStories}~\citep{eldan2023tinystories}.
For language pretraining, we use approximately 20 natural-language
tokens per parameter, with budgets matched across initialization
conditions.
At each transfer boundary, we retain only attention, MLP, and
layer-normalization parameters. Token embeddings, positional embeddings,
and the output head are randomly reinitialized, and optimizer state
is reset.
Training configurations appear in
Appendices~\ref{app:controlled_procedural_training}
and~\ref{app:controlled_standard_pretraining}.
\end{comment}

\section{Specific Improvements Beyond Language Modeling}
\label{sec:evaluating}
\vspace{-2pt}

In this section, we evaluate the effect of
initial training with \textsc{Stack}
on language modeling,
on QA benchmarks with aggregate and question-type performance,
and finally on the synthetic \textsc{Depo} task.

\vspace{-3pt}
\subsection{Improvements in Multi-Hop Question Answering}
\label{sec:real_world}
\vspace{-2pt}

We evaluate SmolLM-135M and SmolLM-360M models
trained on \textsc{Stack} then on \textsc{FineWeb-Edu}
against baselines trained solely on \textsc{FineWeb-Edu}.
These models are then fine-tuned on either \textsc{MuSiQue}, \textsc{HotpotQA}, or \textsc{2WikiMultiHopQA}.
We run this pipeline for 3 pre-training seeds and 3 fine-tuning seeds, i.e.\ evaluating 9 models in total.

\begin{figure}[H]
    \centering
    \vspace{-12pt}
    \includegraphics[width=0.55\linewidth]{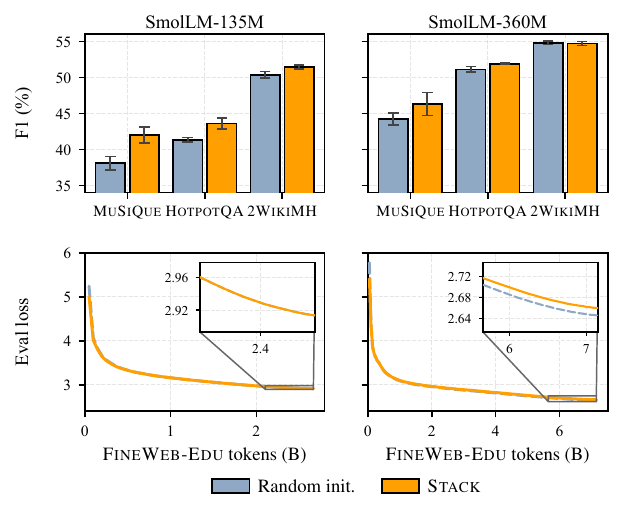}
    \vspace{-13pt}
    \caption{\textbf{\textsc{Stack} improves multi-hop QA without
    improving language modeling.}
    We compare
    \bluetext{language-only pretraining} with
    \orangetext{\textsc{Stack} followed by language pretraining}
    at two model sizes (SmolLM-135M and SmolLM-360M).
    \textbf{(Top)} Token-level \f~after benchmark-specific fine-tuning.
    \textbf{(Bottom)} Validation loss during the pretraining phase on \textsc{FineWeb-Edu}.}
    \label{fig:real_world}
    \vspace{-16pt}
\end{figure}

Figure~\ref{fig:real_world} shows that our models
improve the mean \f~on \textsc{MuSiQue}
and \textsc{HotpotQA} at both sizes, and on
\textsc{2WikiMultihopQA} at 135M.
The result on
\textsc{2WikiMultihopQA} at 360M is largely unchanged.
The largest gain is 3.9 \f~points on \textsc{MuSiQue} with SmolLM-135M.
These improvements occur despite comparable language-modeling
performance at 135M and even slightly worse at 360M
(Figure~\ref{fig:real_world}, bottom).
The initial training on abstract data
therefore shapes the capabilities in the model
in a way that only becomes apparent
after fine-tuning on particular tasks.

\vspace{-2pt}
\subsection{The Gains Concentrate on Sequential Compositions}
\label{sec:reasoning_structure}
\vspace{-2pt}

We now run a finer-grained evaluation
using the annotations of question types in the benchmarks.
These categories specify the type of
reasoning structure necessary to solve each question.
The \textbf{chain}-dominated categories
involve passing intermediate answers
across sequential dependent steps,
while the \textbf{merge}- and \textbf{comparison}-based
categories
involve combining or comparing information from separate branches.
See Appendix~\ref{app:natural_language_evaluation}
for details on the grouping of question types.

\begin{figure}[h]
    \centering
    \vspace{-11pt}
    \includegraphics[width=0.8\linewidth]{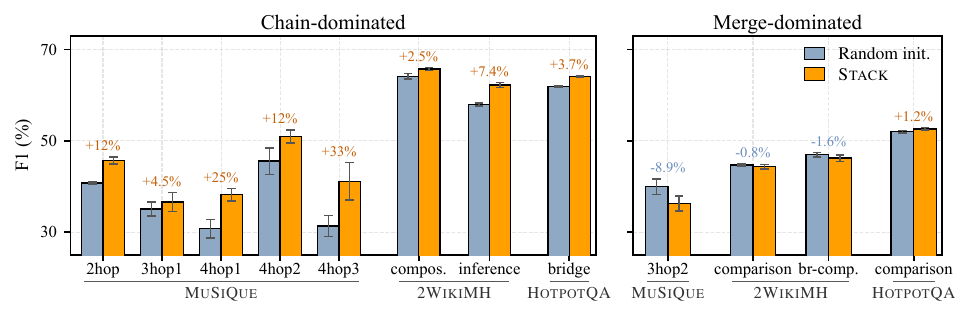}
    \vspace{-9pt}
    \caption{\textbf{Gains on QA benchmarks concentrate on sequentially composition.}
    Mean token-level \f~
    by question type.
    Models trained on \textsc{Stack} perform consistently better in
    every chain-dominated category,
    not in
    merge- and comparison-based categories.
    Annotations indicate relative differences
    with the baseline trained only on language from standard random initialization (no abstract data).
    }
    \label{fig:reasoning_structure}
    \vspace{-13pt}
\end{figure}

Figure~\ref{fig:reasoning_structure}
reports the mean \f~of our models for each question type.
The \textsc{Stack}-trained models
obtain consistently better performance
on chain-dominated questions,
for all relevant question types in the three benchmarks (left box).
In contrast, there is no clear effect on the other question types
in the merge- and comparison-based categories (right box),
with slight degradation on one question type in \textsc{MuSiQue} (close however to the inter-seed variability of the evaluation).
These results indicate that the improvements
obtained with the \textsc{Stack} data
are not generic.
They instead target specific capabilities
in particular for sequential composition.

\subsection{Faster Learning on a Controlled Multi-Hop Task}
\label{sec:controlled_acquisition}
\vspace{-3pt}

We now use
the synthetic \textsc{Depo} task
to precisely measure
how well and how fast the model learns
multi-hop operations of specific complexity.
We use here a simple
GPT-2-style model with 4 layers
(details in Appendix~\ref{app:controlled_standard_pretraining}).
We compare a baseline pretrained on language
(\textsc{FineWeb} or \textsc{TinyStories})
with ours trained first on \textsc{Stack} before language.
They are then fine-tuned for \textsc{Depo}
on data with 1-to-4-hop queries,
and evaluated on test data of each complexity throughout training.

\begin{figure*}[h]
    \centering
    \vspace{-12pt}
    \begin{tikzpicture}
        \node[inner sep=0] (img) {
            \includegraphics[
                width=.95\linewidth,
                trim=0 0 0 1.5mm,
                clip
            ]{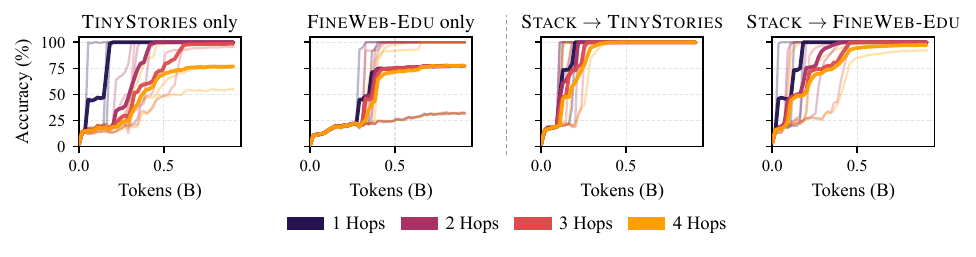}
        };
        \node[font=\small\bfseries, anchor=south]
            at ($(img.north west)!0.305!(img.north east) + (0,0.19mm)$)
            {\footnotesize Baseline: language only};
            %at ($(img.north west)!0.31!(img.north east) + (0,0.19mm)$)
            %{\footnotesize Language Only};
        \node[font=\small\bfseries, anchor=south]
            at ($(img.north west)!0.76!(img.north east) + (0,0.19mm)$)
            {\footnotesize Ours: \textsc{Stack} then language};
            %at ($(img.north west)!0.77!(img.north east) + (0,0.19mm)$)
            %{\footnotesize\textsc{Stack} then language};
    \end{tikzpicture}
    \vspace{-12pt}
    \caption{
    \textbf{(Left)}~On the synthetic \textsc{Depo} task,
    the baseline models with standard pretraining on language
    struggle to perform multi-hop operations.
    \textbf{(Right)}~In comparison, our models exposed to the \textsc{Stack} data
    learn faster and reach near-perfect accuracy across all tested hop counts.
    }
    \label{fig:controlled_acquisition}
    \vspace{-6pt}
\end{figure*}

Figure~\ref{fig:controlled_acquisition} shows that
models exposed to the \textsc{Stack} data
learn \textsc{Depo} faster at every complexity level.
By the end of the fine-tuning,
our models 
reach near-perfect accuracy for all hop counts,
while the language-only models still do not handle the
four-hop queries correctly.
We show in the appendix (Figure~\ref{fig:gpt2_loss_curves})
that this striking improvement on \textsc{Depo}
occurs even though the language modeling validation loss
is similar or even a bit worse
with our model on both language corpora.

\vspace{-8pt}
\takeaway{
Initial training on \textsc{Stack} improves multi-hop QA, with gains
concentrated on sequential composition and long chains,
even when the language modeling capabilities
of the models
appear similar or even slightly worse than the baselines.
}
\vspace{-8pt}

\section{Transfer Depends on Task Structure and Precise Weights}
\label{sec:characterizing}
\vspace{-3pt}

Section~\ref{sec:evaluating} shows that initial
exposure to abstract data can improve performance
after language and task-specific training.
We now try to identify the conditions necessary
for this positive transfer.

%\textsc{Stack}
%training improves multi-hop QA and accelerates the acquisition of
%deep-chain reasoning capabilities despite comparable or worse
%language-modeling performance.
%We now examine how these benefits depend on the structure of the
%abstract task and the weights learned during abstract training.

\vspace{-8pt}
\subsection{Transfer Varies Across Abstract Tasks}
\label{sec:procedural_sweep}
\vspace{-2pt}

\begin{figure}[h]
    \centering
    \vspace{-8pt}
    \includegraphics[width=.85\linewidth]{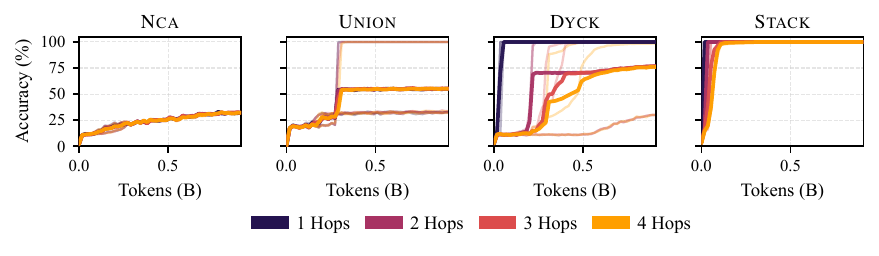}
    \vspace{-8pt}
    \caption{\textbf{Different types of abstract data
    are not equally effective.}
    The test accuracy on \textsc{Depo} throughout training
    is best for \textsc{Stack}.
    See Figure~\ref{fig:depo-search-full} for %comparisons with
    additional options from prior work.
    %the existing literature.
    }
    \label{fig:procedural_sweep}
    %\vspace{-10pt}
\end{figure}

We compare \textsc{Stack} with other
types of abstract data from prior work
(details in Appendix~\ref{app:procedural_tasks}).
We use a minimal setup with direct transfer to \textsc{Depo}
and no intervening language pretraining
(details in Appendix~\ref{app:controlled_procedural_training}).
We observe substantial differences in
Figure~\ref{fig:procedural_sweep}.
%that the effect varies substantially.
%across different types of abstract data.
%Under the evaluated configurations,
\textsc{Stack} enables the fastest learning
to near-perfect accuracy. %on four-hop test cases.
Other options converge more slowly or fail to solve the
3- and 4-hop  test cases.
This highlights the specificity of the effects,
but does not mean that \textsc{Stack}
is universally better: prior work already 
showed that different types of abstract data can induce different capabilities \citep{shinnick2025transformers}.
See Figure~\ref{fig:depo-search-full} in the appendix
for a full comparison of 12 types of data.

%\clearpage

%\vspace{-2pt}
%\subsection{Memory Access and Output Order Affect Transfer}
\label{sec:structural_transfer}
%\vspace{-2pt}

\begin{figure}[h]
    \vspace{-1pt}
    %\centering
%\includegraphics[width=0.6\linewidth]{figures/paper/ablation_structural_stacked.pdf}
\raisebox{14.8pt}{\includegraphics[width=0.499\linewidth,trim=9.5pt 115pt 0 0, clip]{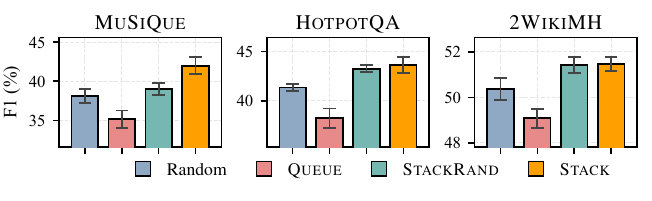}}~%
\includegraphics[width=0.499\linewidth,trim=5pt 0 0 95pt, clip]{figures/paper/ablation_structural_stacked.pdf}
    \vspace{-20pt}
    \caption{
    \textbf{Close variants of the abstract data perform much worse.}
    Replacing the simulated stack with a queue
    is worse than even a random initialization
    on QA benchmarks (\textbf{left}, token-level \f~after language pretraining)
    and on \textsc{Depo} (\textbf{right},
    accuracy after direct transfer).
    %w/o a language phase).
    The \textsc{StackRand} variant
    is closer to the original data and performs accordingly better,
    but still worse  than \textsc{Stack}.}
    \label{fig:stack_computation}
    \vspace{-3pt}
\end{figure}

\begin{minipage}{\linewidth}
    \begin{minipage}[t]{0.56\linewidth}
        \normalsize
        We now evaluate the variants of \textsc{Stack}
        introduced in Section~\ref{sec:procedural_pretraining}:
        \textsc{Queue} and \textsc{StackRand}.
        We evaluate them
        on natural-language multi-hop QA after language pretraining,
        and on \textsc{Depo} through direct transfer.
        \vspace{4pt}
        
        Figure~\ref{fig:stack_computation}
        shows that \textsc{Queue}
        performs poorly in all situations.
        \textsc{StackRand} retains substantial benefits of \textsc{Stack}:
        their performance is close on %QA benchmarks,
        \textsc{HotpotQA} and \textsc{2WikiMultiHopQA},
        but \textsc{StackRand} is a bit worse on \textsc{MuSiQue}
        and learns more slowly on \textsc{Depo}.
        These differences occur again despite
        similar validation loss during language pretraining
        \mbox{(Figure~\ref{fig:structural_lm_loss})}.
        \vspace{4pt}
        
        The precise structure of the abstract task therefore affects
        downstream transfer in ways that the language-modeling loss
        does not capture.
    \end{minipage}\hspace{0.03\linewidth}%
    \begin{minipage}[t]{0.41\linewidth}
        \vspace{-12pt}
        \centering
        \includegraphics[width=.90\linewidth]{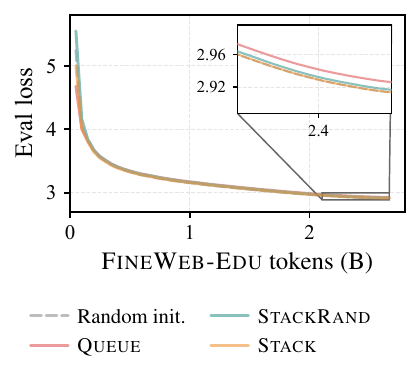}
        \vspace{-10pt}
        \captionof{figure}{
        Similar loss during
        \textsc{FineWeb}-\textsc{Edu} pretraining
        without and with prior exposure to variants of \textsc{Stack}.
        %\textbf{Structural variants yield comparable
        %language-modeling performance.}
        %Validation loss during \textsc{FineWeb-Edu} pretraining from
        %random initialization or after training on \textsc{Stack},
        %\textsc{Queue} (FIFO), or \textsc{StackRand} (shuffled output).
        %Similar validation loss accompanies different QA outcomes
        %(Figure~\ref{fig:stack_computation}).
        }
        \label{fig:structural_lm_loss}
        \end{minipage}
\end{minipage}

\subsection{Adjusting Weight Magnitudes Does Not Reproduce the Full Benefits}
\label{sec:weight_structure}
%\vspace{-1pt}

We test whether the benefits of training on \textsc{Stack}
can be explained by a simple shift in weight magnitudes
facilitating the optimization.
We evaluate two interventions:
after training on \textsc{Stack},
the \textit{Gaussian init.}\ fills
each weight tensor with Gaussian-distributed values
of identical (per-tensor) mean and variance.
The \textit{Shuffled} intervention
instead shuffles values within tensors,
randomizing structure while preserving the exact distribution.
We evaluate them
on \textsc{Depo}
with direct transfer.
%as in previous sections (direct transfer).

\begin{figure}[h]
    \centering
    \vspace{-3pt}
    \includegraphics[width=.95\linewidth]{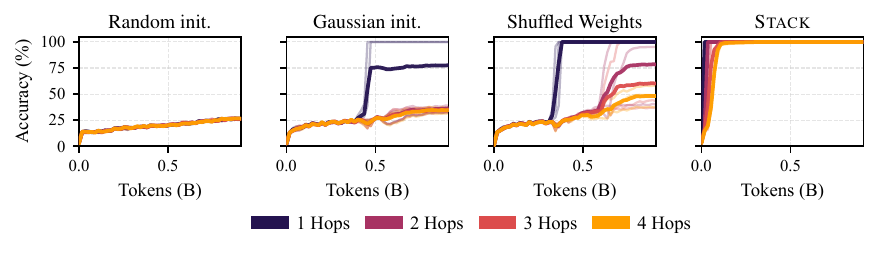}
    \vspace{-12pt}
    \caption{\textbf{Matching weight magnitudes does not reproduce
    \textsc{Stack}'s %full transfer
    benefits.}
    The test accuracy on \textsc{Depo}
    improves slightly (center left/right)
    with random weights whose magnitude
    matches those trained on \textsc{Stack},
    but they do not reach the near-perfect performance
    of the intact weights (right).}
    \label{fig:weight_ablations}
    \vspace{-2pt}
\end{figure}

Figure~\ref{fig:weight_ablations} shows that
both interventions
perform significantly worse than the 
intact \textsc{Stack}-trained weights, particularly
on longer chains.
They retain some benefit over a random initialization,
but neither reproduces \textsc{Stack}'s rapid learning
and near-perfect accuracy.
This means that the information learned from abstract data
is more complex than simple weight magnitudes.

\takeaway{
The benefits of training on abstract data
depend on precise structure within the data,
and the resulting weights
capture more
than a simple adjustment of their magnitudes.
%The structure of the abstract task affects downstream transfer
%despite comparable language-modeling performance.
%Preserving per-tensor weight distributions alone does not reproduce
%\textsc{Stack}'s full benefit for acquiring deep-chain reasoning
%capabilities.
}

\clearpage

 %\section{Early Procedural Training Produces Lasting Gains}
\section{Early Exposure to Abstract Data Produces Lasting Gains}
\label{sec:training_trajectory}

We examine how the timing of abstract data
in the training pipeline impacts its effects,
comparing its use before, during, and after
standard language pretraining.
We then evaluate whether its benefits
persist with increasing amounts of
language pretraining by performing SFT
from various checkpoints of the pretraining trajectory.

\vspace{-3pt}
\subsection{Initial Exposure Produces the Largest Gains}
\label{sec:insertion_timing}
\vspace{-1pt}

\par\vspace{6pt}\noindent
\begin{minipage}{\linewidth}
    \begin{minipage}[t]{0.60\linewidth}
    Using SmolLM-135M, we insert
    a \textsc{Stack} training phase
    at 0\%, 25\%, 50\%, 75\%, or 100\% of a fixed
    \textsc{FineWeb-Edu} token budget.
    \textsc{Stack} uses a separate optimizer,
    such that the language pretraining
    and its learning-rate schedule pause during intermediate
    insertions and then resume.
    We eventually fine-tune each model
    on one of the three QA benchmarks
    %using the same protocol
    and report the mean change in \f~%
    relative to language-only pretraining.
    \vspace{14pt}

    Figure~\ref{fig:stack_insertion_timing} shows that
    the \emph{initial} \textsc{Stack} training
    produces the largest gains.
    %among the tested single-phase schedules.
    Inserting \textsc{Stack} training during language pretraining
    yields smaller gains, while training afterward provides
    no mean QA improvement, even with the best learning rate
    tested (Table~\ref{tab:post_language_lr_sweep}).
    Using \textsc{Stack} later in training therefore
    does not reproduce the gains from initial
    \textsc{Stack} training.
    \end{minipage}\hfill%
    \begin{minipage}[t]{0.35\linewidth}
        \vspace{-13.5pt}
        \centering
        \includegraphics[width=\linewidth]
            {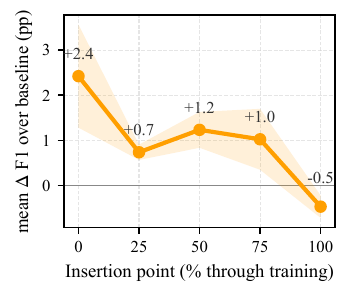}
        \setlength{\abovecaptionskip}{-10pt}
        \captionof{figure}{\textbf{The \emph{initial} training
        on \textsc{Stack} produces the largest gain.}
        Mean change in \f~over the three QA benchmarks relative to
        language-only pretraining.}
        \label{fig:stack_insertion_timing}
    \end{minipage}
\end{minipage}
\par\vspace{6pt}

\begin{minipage}{\linewidth}
    \begin{minipage}[t]{0.41\linewidth}
        \textbf{Controlled training order.}
        We test whether the same timing effect holds for deep-chain
        reasoning capabilities using the controlled setting from
        Section~\ref{sec:controlled_acquisition}.
        For both \textsc{TinyStories} and \textsc{FineWeb-Edu},
        we compare \textsc{Stack} training before and after language
        pretraining, matching abstract- and language-training budgets
        and parameter-transfer protocols between these two settings.
        %Language-only pretraining provides a baseline.
        On both corpora, the initial \textsc{Stack} training
        accelerates the learning on four-hop \textsc{Depo}.
        Training \emph{afterward} on \textsc{Stack} instead
        reduces the final four-hop accuracy below
        the language-only baseline
        (Figure~\ref{fig:depo_training_order}).
    \end{minipage}\hfill%
    \begin{minipage}[t]{0.55\linewidth}
        \vspace{-13pt}
        \centering
        \includegraphics[width=\linewidth,trim=3pt 0 10pt 0, clip]
            {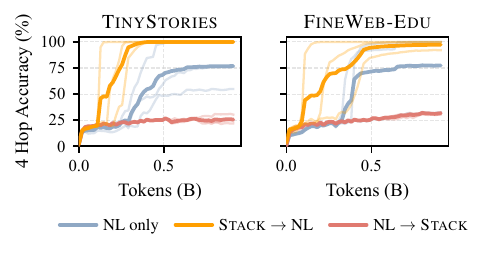}
        \setlength{\abovecaptionskip}{-6pt}
    \captionof{figure}{\textbf{Training order affects deep-chain
    reasoning acquisition.}
    Four-hop \textsc{Depo} learning curves after pretraining on
    natural language (NL) only,
    \textsc{Stack}\,$\rightarrow$\,language, or
    language\,$\rightarrow$\,\textsc{Stack}.}
    \label{fig:depo_training_order}
    \end{minipage}
\end{minipage}

\vspace{8pt}

\begin{minipage}{\linewidth}
    \begin{minipage}[t]{0.57\linewidth}
        %\textbf{Additional training.}
        \textbf{Repeated exposure.}
        We also test whether further training on \textsc{Stack}
        improves on initial training alone.
        Starting with initial \textsc{Stack} training, we add
        three further phases at 25\%, 50\%, and 75\% of language
        pretraining (Figure~\ref{fig:excursion}).
        This schedule increases the amount of abstract training.
        Compared with the initial training alone, it improves mean
        \f~on \textsc{HotpotQA} and \textsc{2WikiMultiHopQA},
        but reduces it on \textsc{MuSiQue}.
        More training on
        this abstract data is therefore not necessarily better.
    \end{minipage}\hfill%
    \begin{minipage}[t]{0.35\linewidth}
        \vspace{-14pt}
        \centering
        \includegraphics[width=.99\linewidth,trim=0 0 -16pt 0, clip]
            {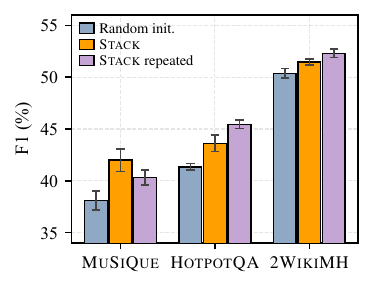}
        \setlength{\abovecaptionskip}{-10pt}
        \captionof{figure}{{Repeated exposure to
        abstract data %\textsc{Stack} data
        has mixed effects.}}
        \label{fig:excursion}
    \end{minipage}
\end{minipage}

\vspace{-8pt}
\subsection{Benefits Persist Throughout Language Pretraining}
\label{sec:persistence}

Finally, we test whether initial \textsc{Stack} training continues
to benefit subsequent natural language multihop QA as language pretraining progresses.
At matched language-token counts, we separately fine-tune checkpoints
from \textsc{Stack}-initialized and randomly initialized SmolLM-135M
models on each QA benchmark, using the same downstream protocol.

Figure~\ref{fig:learning_dynamics} shows that
models given initial \textsc{Stack} training achieve higher %mean
\f~after fine-tuning at every evaluated checkpoint.
%, on all three benchmarks.
Initial abstract training therefore continues to support subsequent
multi-hop QA learning through billions of language-pretraining tokens.

\vspace{7pt}
\begin{minipage}{\linewidth}
    \centering
    \includegraphics[width=0.96\linewidth,trim=0 0 -8pt 0, clip]
        {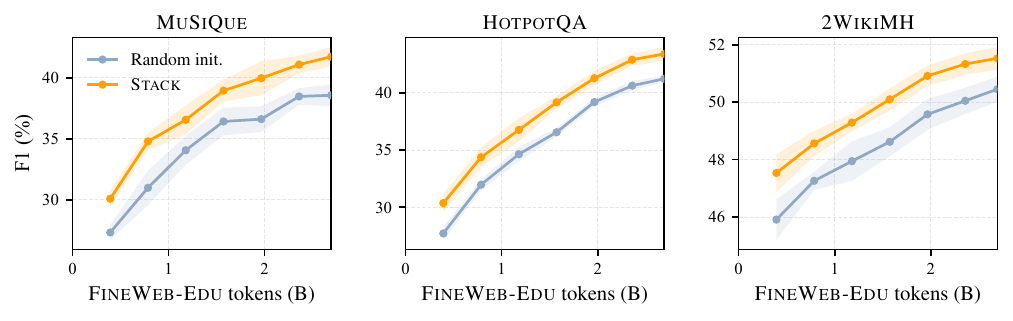}
    \setlength{\abovecaptionskip}{4pt}
    \captionof{figure}{\textbf{The benefits of initial
    \textsc{Stack} training persist throughout language pretraining.}
    Mean token-level \f~after separately fine-tuning successive
    SmolLM-135M checkpoints on each natural language multihop QA benchmark.
    % At matched language-token counts, models given initial
    % \textsc{Stack} training achieve higher mean \f~at every
    % evaluated checkpoint on all three benchmarks.}
    }
    \label{fig:learning_dynamics}
\end{minipage}
\vspace{5pt}

\takeaway{
\emph{Initial} training on \textsc{Stack} produces the largest gains
%mean QA 
among tested single-phase schedules,
while doing so \emph{after} language pretraining provides
no benefit at all. %on the evaluated downstream tasks.
The gains moreover persist through billions of tokens of language pretraining.
}
\vspace{5pt}

\section{Discussion}
\vspace{1pt}

\textbf{Abstract data is best used first.}
Prior work showed that pretraining on abstract data
can improve subsequent language modeling.
But it did not establish whether a dedicated initial phase is necessary.
We directly compared exposure to %the same
abstract data
before, during, and after language.
The benefits are largest when used %abstract training comes
first,
smaller when %it is
mixed with language,
and disappear when introduced afterwards.
The early benefits also persist through billions of %subsequent
language tokens.
This suggests that the abstract data
%meaningfully
impacts the whole training trajectory.
The fact that the benefits cannot be obtained
through post-training
suggests that abstract data instills capabilities
not present in baseline models.
%that our baseline models lack.

\vspace{1pt}
\textbf{The improvements are specific.}
We showed that the abstract data
does not merely provide a generic improvement
in language modeling.
%of the optimization.
In the case of our
\textsc{Stack}
%stack-manipulation
task,
the improvements concentrate
on multi-hop reasoning
and long sequential chains.
The specifics of the abstract data are also critical:
a matched queue-manipulation task does not produce the same benefits.
Future work on abstract data should therefore
seek to identify and align %specific
shared structures
between abstract and target data.

\vspace{1pt}
\textbf{Limitations.}
We study only one type of abstract data and relatively small models.
Other forms of abstract data
e.g.\ from cellular automata \citep{lee2026training}
or random recurrent networks \citep{guo2026synthetic}
may have different effects.
Our results highlight the importance of capability-specific evaluations: improvements may be invisible
in aggregate metrics like perplexity,
but substantial for particular skills.
Our experiments also complement prior work
that already evaluated scalability to larger models~\citep{jiang2026proceduralpretraining}
and longer training~\citep{cheng2026logic}.

%Our experiments are conducted at small scale, but they complement prior studies showing that benefits from abstract pretraining extend to substantially larger models and longer training runs~\citep{jiang2026proceduralpretraining,cheng2026logic}. The present experiments therefore focus on isolating the conditions under which transfer occurs, rather than establishing its scalability.

\vspace{1pt}
\textbf{Future directions.}
We focused on the data and its effects
but we do not explain how it changes the model's parameters.
The simplicity of the abstract data
is an opportunity
%makes it suitable
for a mechanistic analysis of the circuits learned during the initial phase, and how they affect subsequent language training
%There is an opportunity for the tools of mechanistic interpretability
%to analyze how the simple abstract tasks are solved by a model,
%and how the resulting circuits are reused with language
(see e.g.\ \citealt{schiffman2026transformers}).

Another question is whether the benefits of abstract training can be obtained more efficiently.
We showed that the initial phase shapes the optimization trajectory
by modifying the inductive biases at initialization.
If we could characterize these inductive biases more directly
\citep{teney2024neural,teney2026can}
it may be possible to obtain similar benefits
with new architectures or initializations,
rather than an explicit abstract-training phase.
This could instill useful generic structures
into the weights of a model in lieu of the arbitrary, random initializations in use today.
%\clearpage

\newpage
\subsection*{AI use statement}
Generative AI was used solely to assist with grammar and presentation. It did not contribute to the development of the core ideas, experimental design or execution, or analysis
%, or interpretation
of the results.

\subsection*{Reproducibility statement}
We provide detailed descriptions of our methods, experimental settings, and evaluation procedures in the %main paper and
appendix. We will release the code required
to reproduce our experiments upon publication.

\bibliography{references}
\bibliographystyle{iclr2027_conference}

\appendix
\clearpage
\section{Procedural Pretraining Details}
\label{app:procedural_training}

% ------------------------------------------------------------
\subsection{Procedural Tasks}
\label{app:procedural_tasks}

\paragraph{\textsc{Stack}.}
We adopt the \textsc{Stack} task from
\citet{jiang2026proceduralpretraining} and use the authors'
implementation.\footnote{\url{https://github.com/zlshinnick/procedural-pretraining}}
Each example consists of a sequence of push and pop operations over
abstract symbols. A push operation adds a symbol to the top of the
stack, while a pop operation removes the current top element. After
executing the full sequence, the model predicts the remaining stack
contents from top to bottom. The task therefore requires maintaining
and updating an internal state under last-in-first-out (LIFO) access.
% For example,
% \[
% \texttt{push(a), push(b), pop, push(c)}
% \quad\longrightarrow\quad
% \texttt{c, a}.
% \]
% \damien{Not a good description. What is the meaning of the arrow? What corresponds to one token? What is the vocabulary? How is the data generated? What is the loss (NTP, but is there only 1 correct continuation? NO in principle!); where is the loss applied? You don't describe any of the most important points!}
The task contains no natural-language semantics: symbols are arbitrary,
and the target is determined entirely by the sequence of stack
operations. Training configurations for the natural-language and
controlled settings are provided in
Appendices~\ref{app:nl_procedural_training}
and~\ref{app:controlled_procedural_training}, respectively. Table~\ref{tab:stackExample} provides an example.

\paragraph{Structural controls.}
We construct two variants using exactly the same input sequences
as \textsc{Stack}.
The FIFO control changes the interpretation of the pop token:
it removes the oldest element rather than the most recently added
element, and the remaining queue contents are emitted front-to-back. 
% For example,
% \[
% \texttt{push(a), push(b), pop, push(c)}
% \quad\longrightarrow\quad
% \texttt{b, c}.
% \]
The shuffled-output control preserves the original LIFO state updates
but independently shuffles the final stack contents for each example
before emitting them.
% For example,
% \[
% \texttt{push(a), push(b), pop, push(c)}
% \;\xrightarrow{\textsc{Stack}}\;
% \texttt{c, a}
% \;\xrightarrow{\text{shuffle}}\;
% \texttt{a, c}.
% \]
All other data-generation rules follow the original implementation. Table~\ref{tab:stackExample} provides examples.

\paragraph{Alternative procedural tasks.}
In the controlled setting only, we additionally evaluate a broad suite
of procedural tasks spanning sequence transformations, formal languages,
and cellular automata. Following
\citet{jiang2026proceduralpretraining}, we use
\textsc{Identity}, which copies the input sequence;
\textsc{Reverse}, which outputs it in reverse order;
\textsc{Delete}, which removes a specified token;
\textsc{Set}, which removes duplicate tokens;
\textsc{Sort}, which outputs the sequence in sorted order; and
\textsc{Union}, which combines two sequences while removing duplicates.
We use neural cellular automata (NCA) following
\citet{lee2026training}, with configurations of increasing complexity.
Finally, following the formal-language setup of
\citet{hu2025between}, we use \textsc{Dyck}, which models nested bracket
structure, and \textsc{Dyck-Shuffle}, a matched non-nested variant.
Exact controlled-training configurations are provided in
Appendix~\ref{app:controlled_procedural_training}.

% ------------------------------------------------------------
\subsection{Natural-Language Procedural Training}
\label{app:nl_procedural_training}

\paragraph{Data construction and objective.}
Stage-(1) procedural pretraining uses the \textsc{Stack} task described
in Appendix~\ref{app:procedural_tasks}.
Each example contains 128 input tokens, followed by a separator and
the final stack contents read from top to bottom.
We apply the language-modeling loss only to output tokens,
masking the input and separator.

\paragraph{Vocabulary and transfer.}
For each architecture, we construct the procedural task using the
model's native vocabulary. The full
procedurally pretrained checkpoint can therefore be transferred directly
into standard pretraining, including token embeddings and the output
head, without resizing or reinitialization.

\paragraph{Optimization.}
We train for 5,000 steps with effective batch size 64. Hidden attention
and MLP parameters are optimized with Muon using learning rate
$4\times10^{-3}$, momentum $0.95$, and weight decay $0.05$.
Embeddings, the output head, and remaining parameters use an auxiliary
AdamW optimizer with learning rate $10^{-4}$; weight decay is applied
to matrix parameters but not one-dimensional parameters. We use 250 warm-up steps followed by
cosine decay to 10\% of the peak learning rate.

% ------------------------------------------------------------
\subsection{Controlled Procedural Training}
\label{app:controlled_procedural_training}

\paragraph{Model and optimization.}
All controlled procedural experiments use a 4-layer, 4-head,
512-dimensional GPT-2 transformer ($\sim$12.7M parameters).
We use AdamW throughout. Table~\ref{tab:procedural_task_configs}
reports the exact data and optimization configurations for the
procedural-task sweep.

\begin{table*}[t]
\centering
\scriptsize
\setlength{\tabcolsep}{5pt}
\renewcommand{\arraystretch}{1.12}

\begin{tabular}{lcccl}
\toprule
\textbf{Task}
& \textbf{Seq. len.}
& \textbf{Vocab.}
& \textbf{Optimization}
& \textbf{Task-specific setting} \\
\midrule

\multicolumn{5}{l}{\textit{Procedural tasks from
\citet{jiang2026proceduralpretraining}}} \\

\textsc{Stack}
& 128 & 102
& 10k steps, eff. bs 64, $5{\times}10^{-4}$ const.
& push/pop; emit stack top-first \\

\textsc{Identity}
& 128 & 102
& 10k steps, eff. bs 64, $5{\times}10^{-4}$ const.
& copy input \\

\textsc{Reverse}
& 128 & 102
& 10k steps, eff. bs 64, $5{\times}10^{-4}$ const.
& reverse input \\

\textsc{Delete}
& 128 & 102
& 10k steps, eff. bs 64, $5{\times}10^{-4}$ const.
& remove specified token \\

\textsc{Set}
& 128 & 102
& 10k steps, eff. bs 64, $5{\times}10^{-4}$ const.
& emit unique elements \\

\textsc{Sort}
& 128 & 102
& 10k steps, eff. bs 64, $5{\times}10^{-4}$ const.
& sort input \\

\textsc{Union}
& 128 & 102
& 10k steps, eff. bs 64, $5{\times}10^{-4}$ const.
& deduplicated union \\

\midrule

\multicolumn{5}{l}{\textit{Neural cellular automata following
\citet{lee2026training}}} \\

NCA-Low
& 1024 & 64,000
& 5k steps, eff. bs 32, $10^{-4}$ cos.
& complexity $0.10$--$0.25$ \\

NCA-Medium
& 1024 & 64,000
& 5k steps, eff. bs 32, $10^{-4}$ cos.
& complexity $0.25$--$0.50$ \\

NCA-High
& 1024 & 64,000
& 5k steps, eff. bs 32, $10^{-4}$ cos.
& complexity $0.50$--$1.00$ \\

\midrule

\multicolumn{5}{l}{\textit{Formal languages following
\citet{hu2025between}}} \\

\textsc{Dyck}
& 2048 & 129
& 500 steps, eff. bs 32, $5{\times}10^{-4}$ const.
& $k=64$ \\

\textsc{Dyck-Shuffle}
& 2048 & 129
& 500 steps, eff. bs 32, $5{\times}10^{-4}$ const.
& $k=64$ \\

\midrule

\multicolumn{5}{l}{\textit{Structural controls introduced in this work}} \\

Shuffled-output \textsc{Stack}
& 128 & 102
& 10k steps, eff. bs 64, $5{\times}10^{-4}$ const.
& LIFO state updates; permuted read-out \\

FIFO
& 128 & 102
& 10k steps, eff. bs 64, $5{\times}10^{-4}$ const.
& pop oldest; emit oldest-first \\

\bottomrule
\end{tabular}

\caption{\textbf{Procedural pretraining configurations used in the
controlled experiments.}
All models are 4-layer, 4-head, 512-dimensional GPT-2 transformers
($\sim$12.7M parameters) trained with AdamW over three seeds. For
tasks adopted from prior work, we follow the original protocols as
closely as possible while adapting them to our controlled model and
transfer setting. The FIFO and shuffled-output \textsc{Stack} controls
use the same training configuration as \textsc{Stack}.}
\label{tab:procedural_task_configs}
\end{table*}

\paragraph{Training objective.}
For the input--output procedural tasks from
\citet{jiang2026proceduralpretraining}, the language-modeling loss is
applied only to output tokens. The NCA tasks follow the next-token
prediction setup of \citet{lee2026training}, while
\textsc{Dyck} and \textsc{Dyck-Shuffle} use full-sequence
language-modeling loss following \citet{hu2025between}.

\paragraph{Transfer setup.}
Following \citet{jiang2026proceduralpretraining, lee2026training}, we transfer the
transformer blocks from the procedural checkpoint into subsequent
training stages while reinitializing vocabulary-specific parameters
where required. For direct transfer to DEPO, token embeddings,
positional embeddings, and the language-modeling head are freshly
initialized. All parameters remain trainable after transfer.

% ============================================================
\clearpage
\section{Standard Pretraining Details}
\label{app:nl_training}

% ------------------------------------------------------------
\subsection{Natural-Language Standard Pretraining}
\label{app:nl_standard_pretraining}

\paragraph{Data.}
We use \textsc{FineWeb-Edu}~\citep{penedo2024fineweb} with the
49,152-token SmolLM tokenizer. EOS-terminated documents are concatenated
into 2,048-token blocks, discarding incomplete blocks and allowing
cross-document attention. Each run uses blocks without repetition.

\paragraph{Models and training budgets.}
We train SmolLM-135M and SmolLM-360M with an effective batch size
of 96 sequences and approximately Chinchilla-optimal budgets of
20 tokens per parameter.
Table~\ref{tab:nl_pretraining_budgets} summarizes the budgets.

\begin{table}[h]
    \centering
    \small
    \caption{\textbf{Language-pretraining budgets.}
    All runs use 2,048-token sequences and an effective batch size
    of 96 sequences.}
    \label{tab:nl_pretraining_budgets}
    \begin{tabular}{lrr}
        \toprule
        Model & Language tokens & Optimization steps \\
        \midrule
        SmolLM-135M & 2.69B & 13,701 \\
        SmolLM-360M & 7.20B & 36,621 \\
        \bottomrule
    \end{tabular}
\end{table}

\paragraph{Optimization and hyperparameter selection.}
We use Muon for hidden attention and MLP matrices, with auxiliary
AdamW for embeddings, the output head, and remaining parameters.
Muon uses Nesterov momentum with coefficient $0.95$.
Auxiliary AdamW uses $\beta_1=0.9$, $\beta_2=0.95$,
and $\epsilon=10^{-10}$.
To establish a tuned baseline, we perform a hyperparameter grid
search on randomly initialized SmolLM-135M, training every
configuration for the full Chinchilla-optimal budget of 2.69B tokens.
We select the configuration with the lowest validation loss:
Muon learning rate $0.016$, weight decay $0.01$, and auxiliary
AdamW learning rate $3\times10^{-3}$
(Table~\ref{tab:muon_hyperparameters}).
We use these settings for all SmolLM language-pretraining runs.
Both optimizers use linear warm-up over the first 5\% of training
steps, followed by cosine decay to 10\% of their respective peak
learning rates.

\begin{table}[h]
    \centering
    \small
    \setlength{\tabcolsep}{6pt}
    \renewcommand{\arraystretch}{1.1}
    \caption{\textbf{Muon hyperparameter search on SmolLM-135M.}
    Each configuration is trained for 2.69B \textsc{FineWeb-Edu}
    tokens. The selected configuration has the lowest validation
    loss (in bold).}
    \label{tab:muon_hyperparameters}
    \begin{tabular}{cccc}
        \toprule
        Muon LR & Weight decay & Auxiliary AdamW LR
        & Validation loss $\downarrow$ \\
        \midrule
        $0.016$ & $0.01$ & $3\times10^{-3}$ & \textbf{2.9144} \\
        $0.020$ & $0.01$ & $3\times10^{-3}$ & 2.9150 \\
        $0.010$ & $0.01$ & $3\times10^{-3}$ & 2.9225 \\
        $0.008$ & $0.01$ & $3\times10^{-3}$ & 2.9275 \\
        $0.008$ & $0.10$ & $3\times10^{-3}$ & 2.9361 \\
        $0.010$ & $0.10$ & $3\times10^{-3}$ & 2.9430 \\
        $0.008$ & $0.10$ & $3\times10^{-4}$ & 2.9608 \\
        $0.010$ & $0.01$ & $3\times10^{-4}$ & 2.9662 \\
        $0.010$ & $0.10$ & $3\times10^{-4}$ & 2.9674 \\
        $0.016$ & $0.01$ & $3\times10^{-4}$ & 2.9687 \\
        $0.008$ & $0.01$ & $3\times10^{-4}$ & 2.9689 \\
        $0.016$ & $0.10$ & $3\times10^{-3}$ & 2.9695 \\
        $0.020$ & $0.01$ & $3\times10^{-4}$ & 2.9756 \\
        $0.020$ & $0.10$ & $3\times10^{-3}$ & 2.9884 \\
        $0.016$ & $0.10$ & $3\times10^{-4}$ & 2.9982 \\
        $0.020$ & $0.10$ & $3\times10^{-4}$ & 3.0195 \\
        \bottomrule
    \end{tabular}
\end{table}

\paragraph{Initialization and controls.}
The baseline starts from random initialization, whereas the
procedural condition retains the full stage-(1) model, including
token embeddings and the language-modeling head.
Both conditions use the same stage-(2) training data, token budget,
and optimization settings.

\paragraph{Random seeds.}
For SmolLM-135M, we use three language-pretraining seeds per setting.
Each procedural run starts from an independently trained \textsc{Stack}
checkpoint paired with a distinct language-pretraining seed.
For SmolLM-360M we use one pretraining seed due to computational cost.
Downstream fine-tuning seeds are specified in
Appendix~\ref{app:natural_language_evaluation}.

% ------------------------------------------------------------
\subsection{Controlled Standard Pretraining}
\label{app:controlled_standard_pretraining}

\paragraph{Data and model.}
We perform controlled language pretraining on \textsc{TinyStories}
and \textsc{FineWeb-Edu}.
For \textsc{TinyStories}, we use
\texttt{noanabeshima/TinyStoriesV2}.
For both datasets, we use a 10,000-token BPE tokenizer.
Both settings use a GPT-2 transformer with four layers, four
attention heads, hidden dimension 512, dropout 0.1, and tied
input and output embeddings, with approximately 12.6M
non-embedding parameters.

\paragraph{Optimization and budget.}
We use AdamW with learning rate $5\times10^{-4}$,
$\beta=(0.9,0.999)$, $\epsilon=10^{-8}$, and no weight decay.
The effective batch size is 32 sequences and sequence length is 2048.
We train for 5,731 optimization steps, using 10\% linear warm-up
followed by cosine decay to $5\times10^{-5}$.
We target a Chinchilla-optimal budget of approximately 20 language tokens
per parameter, using identical language-training recipes for random and
procedural initializations within each corpus.

% ------------------------------------------------------------
\subsection{Natural-Language Multi-Hop Evaluation}
\label{app:natural_language_evaluation}

\paragraph{Benchmarks and context.}
We evaluate on \textsc{MuSiQue} using gold supporting paragraphs,
and on \textsc{HotpotQA} and \textsc{2WikiMultihopQA} using all
provided paragraphs, including distractors.
Models receive the context and question and are trained to produce
short answers without reasoning traces.

\paragraph{Fine-tuning.}
We fine-tune all parameters for one epoch using AdamW with learning
rate $3\times10^{-4}$, weight decay $0.01$, and effective batch size 32.
We use 3\% linear warm-up followed by cosine decay, with loss applied
only to answer and EOS tokens.

\paragraph{Context-length filtering.}
We exclude examples exceeding 2,048 tokens, including answer and EOS
tokens during training or a 64-token generation allowance at evaluation.
Retained subsets are identical across model sizes and initializations
(Table~\ref{tab:qa_dataset_counts}).

\begin{table}[t]
    \centering
    \small
    \caption{\textbf{Retained natural language QA benchmark examples after context-length filtering.}
    Counts are retained/original examples using the SmolLM tokenizer.}
    \label{tab:qa_dataset_counts}
    \begin{tabular}{lrr}
        \toprule
        Benchmark & Training & Development \\
        \midrule
        \textsc{MuSiQue} & 19,937/19,938 & 2,417/2,417 \\
        \textsc{HotpotQA} & 18,786/20,000 & 6,813/7,405 \\
        \textsc{2WikiMultihopQA} & 19,674/20,000 & 12,075/12,576 \\
        \bottomrule
    \end{tabular}
\end{table}

\paragraph{Evaluation.}
We evaluate final checkpoints on development splits using greedy
decoding (up to 64 tokens), taking the first output line as the answer.
We report \f~and exact match using the official \textsc{MuSiQue}
evaluation code~\citep{trivedi2022musique}.

\paragraph{Random seeds.}
Each pretrained checkpoint is fine-tuned with three random seeds,
yielding nine runs per setting for SmolLM-135M (three checkpoints)
and three for SmolLM-360M (one checkpoint).
We report means and standard deviations across runs, capturing both
pretraining and fine-tuning variation for SmolLM-135M.

\paragraph{Reasoning structure.}
We classify official question types by the dominant (longest-path)
structure of their composition graphs.
Chain-dominated types pass intermediate answers between dependent hops,
whereas merge-dominated types join independent lookups.

% ------------------------------------------------------------
\subsection{Controlled Multi-Hop Evaluation}
\label{app:depo}

\paragraph{DEPO task.}
Each example contains a directed cycle over 10 distinct entities,
sampled from a vocabulary of 20 single-token entities.
The 10 directed edges are presented in random order, followed by
10 queries, one for each starting entity.
Each query asks for the entity reached after $k$ successive
transitions, with $k$ sampled uniformly from $\{1,2,3,4\}$ during
training.
Loss is applied only to answer tokens.

\paragraph{Example.}
For a cycle beginning $a\rightarrow b\rightarrow c\rightarrow d$,
a query starting at $a$ with $k=3$ has answer $d$.
Edges are encoded as \texttt{SEP a b}, and queries as
\texttt{Q\_k a ANS d}, where \texttt{Q\_k} specifies the hop count.
Each complete example contains 71 tokens.

\paragraph{Model and optimization.}
We use the GPT-2 transformer with four layers, four attention heads,
hidden dimension 512.
We train for 100,000 steps with batch size 128 using AdamW,
learning rate $5\times10^{-4}$, $\beta=(0.9,0.999)$,
$\epsilon=10^{-8}$, and no weight decay.
We use 1,000 linear warm-up steps followed by cosine decay to zero.

\paragraph{Evaluation protocol.}
We evaluate  every 2,000 steps on a fixed set
of 500 examples per hop, giving 5,000 answer predictions for each
$k\in\{1,2,3,4\}$.

\clearpage
\section{Additional Results}
\label{app:additional-results}
\vspace{-5pt}

\begin{figure}[h]
    \centering
    \includegraphics[width=0.8\linewidth]{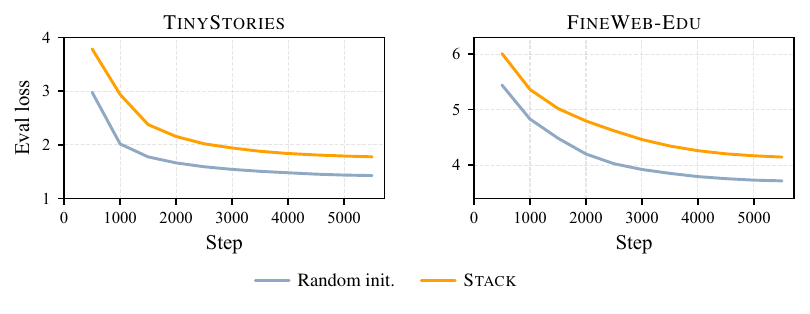}
    \vspace{-10pt}
    \caption{\textbf{Language-modeling validation loss on \textsc{TinyStories} and \textsc{FineWeb-Edu}.}
    \textsc{Stack}-initialized models exhibit higher loss than randomly
    initialized models throughout language pretraining.}
    \label{fig:gpt2_loss_curves} 
    %\vspace{-5pt}
\end{figure}

\begin{figure}[h]
    \centering
    \includegraphics[width=0.9\linewidth]{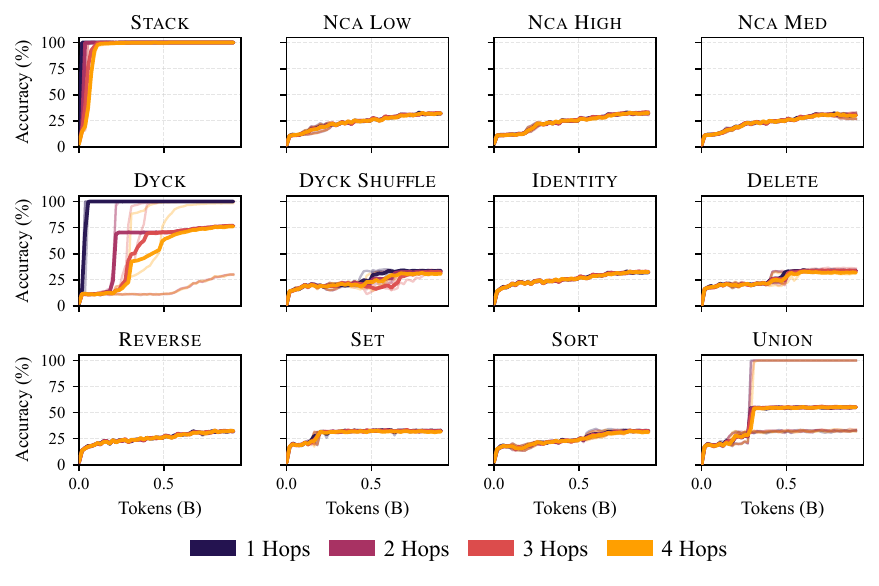}
    \vspace{-8pt}
    \caption{\textbf{Full comparison of abstract data on \textsc{DEPO}.}
    Accuracy by hop count during downstream training after initial
    training on abstract data, without intervening language pretraining.
    \textsc{Stack} enables the fastest learning to near-perfect accuracy
    across all four hop counts; other types of abstract data yield
    slower or less complete learning, particularly on longer chains.}
    \label{fig:depo-search-full}
    %\vspace{-10pt}
\end{figure}

\begin{table}[h]
    \centering
    \caption{\textbf{Learning-rate sweep for post-language
    \textsc{Stack} training.}
    Downstream \f~for SmolLM-135M, using three pretraining seeds
    per configuration. Language-only and initial \textsc{Stack}
    exposure are included as references.}
    \label{tab:post_language_lr_sweep}
    \vspace{-2pt}
    \small
    \begin{tabular}{llrrr}
        \toprule
        Muon LR & Auxiliary LR & MuSiQue & HotpotQA & 2WikiMH \\
        \midrule
        $4\times10^{-3}$ & $10^{-4}$ & 23.57 & 25.07 & 44.96 \\
        $1.3\times10^{-3}$ & $3.3\times10^{-5}$ & 35.97 & 38.66 & 49.26 \\
        $4\times10^{-4}$ & $10^{-5}$ & 37.93 & 40.57 & 50.01 \\
        \midrule
        \multicolumn{2}{l}{Language-only baseline}
        & 38.10 & 41.34 & 50.37 \\
        \multicolumn{2}{l}{\textsc{Stack}\,$\rightarrow$\,\textsc{FineWeb-Edu}}
        & 41.99 & 43.54 & 51.24 \\
        \bottomrule
    \end{tabular}
    \vspace{-10pt}
\end{table}

\end{document}